\documentclass[conference]{IEEEtran}
\IEEEoverridecommandlockouts
\usepackage{cite}
\usepackage{amsmath,amssymb,amsfonts}
\usepackage{algorithmic}
\usepackage{graphicx, textcomp, xcolor, soul}
\usepackage{etoolbox}
\usepackage{enumitem}

\usepackage{comment}
\def\BibTeX{{\rm B\kern-.05em{\sc i\kern-.025em b}\kern-.08em
    T\kern-.1667em\lower.7ex\hbox{E}\kern-.125emX}}
\begin{document}

\title{Machine-Learning–Powered Neural Interfaces for Smart Prosthetics and Diagnostics}

\author{
\IEEEauthorblockN{MohammadAli Shaeri\textsuperscript{*}, Jinhan Liu\textsuperscript{*}, Mahsa Shoaran}
\IEEEauthorblockA{\textit{Institutes of Electrical and Micro Engineering and Neuro-X, EPFL} \\
Lausanne, Switzerland \\
\{mohammad.shaeri, jinhan.liu, mahsa.shoaran\}@epfl.ch}
\vspace{-5mm}

\thanks{\textsuperscript{*} These authors contributed equally to this work.}
}

\maketitle
\begin{abstract}
Advanced neural interfaces are transforming applications ranging from neuroscience research to diagnostic tools (for mental state recognition, tremor and seizure detection) as well as prosthetic devices (for motor and communication recovery). By integrating complex functions into miniaturized neural devices, these systems unlock significant opportunities for personalized assistive technologies and adaptive therapeutic interventions. Leveraging high-density neural recordings, on-site signal processing, and machine learning (ML), these interfaces extract critical features, identify disease neuro-markers, and enable accurate, low-latency neural decoding. This integration facilitates real-time interpretation of neural signals, adaptive modulation of brain activity, and efficient control of assistive devices. Moreover, the synergy between neural interfaces and ML has paved the way for self-sufficient, ubiquitous platforms capable of operating in diverse environments with minimal hardware costs and external dependencies.
In this work, we review recent advancements in AI-driven decoding algorithms and energy-efficient System-on-Chip (SoC) platforms for next-generation miniaturized neural devices. These innovations highlight the potential for developing intelligent neural interfaces, addressing critical challenges in scalability, reliability, interpretability, and user adaptability.
\end{abstract}

\begin{IEEEkeywords}
Neural Interfaces, Neuromodulation, Brain-Computer Interfaces (BCI), Machine Learning, System-on-Chip. % Brain Implant Hardware Efficiency
\end{IEEEkeywords}

\vspace{-5mm}
\section{Introduction}
\vspace{-1mm}
Modern neural interfaces can simultaneously record thousands of neural signals \cite{patrick2024state}. This capability generates massive volumes of data that demand efficient processing and compression to extract meaningful insights \cite{Shaeri2015TNSRE, shoaran2014compact, Shaeri2022Xform} (Fig. \ref{fig: NI Categories}(a)). Furthermore, advances in neural interface technology have enabled real-time recording and transmission, paving the way for high-resolution mapping of brain activity \cite{shin2022neuraltree, ding202549}. %huang2022actively, shin2022256
However, the scale and complexity of neural data pose significant challenges in processing, storage, and interpretation, necessitating innovative approaches in both hardware and software.

Machine learning (ML) has emerged as a transformative tool for data analysis, enabling the execution of sophisticated tasks that were previously unattainable. ML algorithms excel at identifying patterns in complex datasets, making them invaluable for applications such as brain-computer interfaces (BCIs) aimed at motor and communication restoration \cite{flesher2021brain, willett2021high, silva2024speech, lorach2023walking}. %willett2023high, metzger2023high, milekovic2023spinal
Additionally, ML facilitates therapeutic applications, including mental state recognition, migraine detection, and seizure prediction \cite{Shoaran2024Intelligent, liu2024neural, Afzal2024REST, zhu2019migraine, yao2021predicting}.
Moreover, brain-triggered neuromodulation systems leverage ML to adaptively modulate brain activity in response to disease-specific neuromarkers, offering personalized and precise therapeutic interventions \cite{shin2022neuraltree, tsai2023seizure, yao2020improved}.

To achieve these advanced functionalities, integrating neural interfaces and ML into miniaturized systems has become a key research focus \cite{shoaran2018energy, yoo2021neural, Shaeri2024MiBMI, YS2024JSSC, shin202216, tsai2023seizure}. Existing works have demonstrated promising results in developing energy-efficient system-on-chip (SoC) platforms for real-time neural signal processing and decoding. However, significant challenges remain in achieving hardware efficiency, ensuring scalability, and maintaining the reliability and interpretability of these systems. Addressing these challenges is essential for advancing next-generation neural interfaces that combine compact designs with robust performance in real-world applications.
%\cite{marblestone2013physical, Shoaran2024Intelligent}

% General Trends, A note on patients, Monkey and patients with motor disabilities, Regression vs classification

% \section{Neural Devices Applications}
\vspace{-1mm}
\section{Neural Interfaces}
% Advancements in therapeutic neural devices are revolutionizing the diagnosis and treatment of neurological and psychiatric disorders, enabling real-time monitoring, adaptive neurostimulation, and personalized interventions. These systems enhance seizure detection, tremor suppression, emotion regulation, and migraine therapy, improving clinical outcomes and patient independence through closed-loop neural decoding and intervention.

\begin{figure}[ht]
    \centering    \includegraphics[width=0.85 \columnwidth]{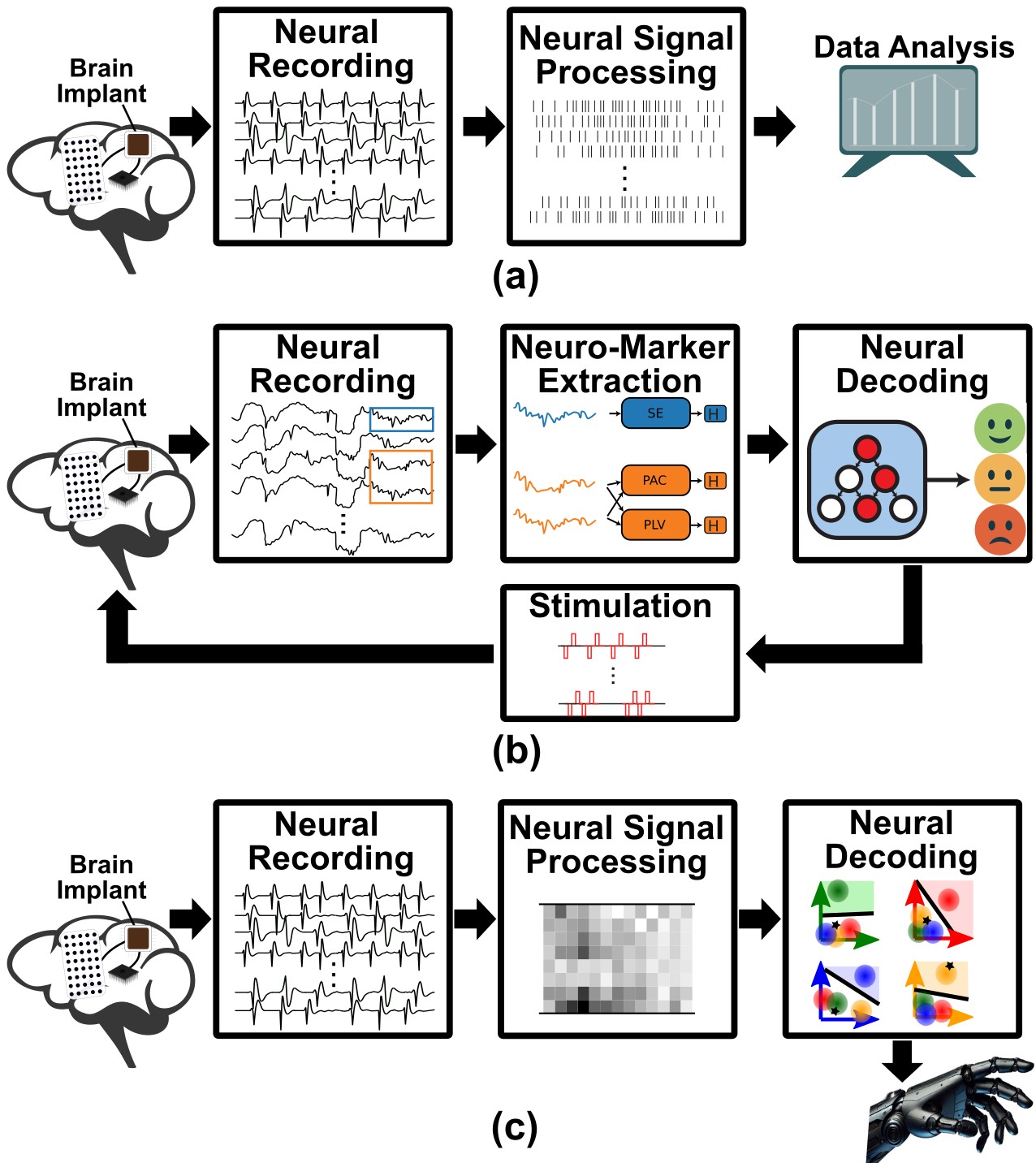}
    \vspace{-10pt}
    \caption{\textbf{Neural Interfaces:} (a) Neural recording interfaces capture high-density neural signals and process them to reduce data rates, or wirelessly transmit them to an external computer for further processing. (b) Therapeutic neural interfaces extract neuro-markers to detect disease-related neurological symptoms or mental states, and may also integrate neurostimulation for functions such as seizure suppression or brain rewiring. (c) Prosthetic neural interfaces use ML to convert brain intention into actionable commands, enabling control of end-effectors like robotic hands.}
    \vspace{-10pt}
    \label{fig: NI Categories}
\end{figure}

\textbf{Diagnostic/Therapeutic Applications}--- ML can empower advanced therapeutic  devices and  transform the diagnosis and treatment of neurological and psychiatric disorders, enabling real-time monitoring, adaptive neurostimulation, and personalized interventions (Fig. \ref{fig: NI Categories}(b)). These systems enhance clinical outcomes and patient independence via seizure/tremor suppression, emotion regulation, and migraine therapy.
%\cite{ramgopal2014seizure-breif}.
% Neural interfaces for epilepsy management integrate real-time seizure detection and closed-loop neuromodulation, delivering targeted electrical stimulation or drug therapy to prevent seizures. By analyzing spatiotemporal neural activity, these devices enhance prediction accuracy and response time, reducing seizure frequency and severity \cite{Afzal2024REST, tsai2023seizure}. Similarly, DBS-based neurostimulation systems for Parkinson’s disease and essential tremor detect and modulate abnormal oscillations, dynamically adjusting stimulation parameters to optimize motor control and treatment efficacy.
By integrating real-time detection of neurological symptoms (e.g., seizures) with closed-loop neuromodulation, these devices deliver targeted stimulation to prevent seizures %by analyzing spatiotemporal neural activity%, improving prediction accuracy and response time
\cite{shin2022neuraltree, tsai2023seizure}.
Similarly, closed-loop neural interfaces can be used to treat tremors in patients with Parkinson's disease by detecting and modulating abnormal oscillations \cite{shin2022neuraltree, yao2020improved}.
%Similarly, DBS-based systems for Parkinson’s disease and tremor adjust stimulation parameters dynamically, optimizing motor control and treatment efficacy.
% For psychiatric and mental health disorders, neural interfaces decode emotional and anxiety-related states from neural signals, enabling real-time monitoring and personalized interventions. These systems analyze neural connectivity and spectral dynamics to assess mood disorders and anxiety, guiding adaptive neurostimulation or biofeedback therapy \cite{song2022eeg, du2020efficient, liu2024neural}. By combining real-time decoding with closed-loop interventions, these devices support precision psychiatry and personalized mental health treatments, advancing neurotechnology for cognitive and emotional well-being. 
Neural interfaces are also being applied in the treatment of psychiatric disorders, for example, in decoding emotional and anxiety-related states, analyzing connectivity and spectral dynamics to guide adaptive neurostimulation and biofeedback therapy \cite{liu2024neural, shin202216, shin202316}. % du2020efficient, ding2023lggnet, song2022eeg, %By combining real-time decoding with closed-loop interventions, these devices advance precision psychiatry and cognitive neurotechnology.

% Implantable neurostimulation devices for migraine management track cortical spreading depression (CSD) patterns, a key biomarker in migraine onset. These systems apply targeted electrical stimulation to interrupt migraine progression, reducing pain intensity and frequency. Adaptive algorithms refine stimulation delivery based on individual neural responses, making the treatment highly personalized and minimally invasive.

\textbf{Prosthetic/Assistive Applications}---
Prosthetic neural interfaces employ ML to decode brain signals in order to drive a wide range of effectors, enabling brain control of prosthetic limbs and communication devices for individuals with disabilities (Fig. \ref{fig: NI Categories}(c)). Such interfaces provide critical functionalities such as artificial limb movement \cite{aflalo2015decoding}, % hochberg2012reach, collinger2013high
locomotion assistance through wheelchair operation \cite{libedinsky2016independent} and direct activation of natural limbs via neuromuscular \cite{ajiboye2017restoration-breif} %bouton2016restoring
or spinal cord stimulation \cite{lorach2023walking}, %milekovic2023spinal, courtine2019spinal, 
facilitating the restoration of walking and hand movements.
%gait, reach-and-grasp, 
Also, neural prostheses empower individuals to perform daily tasks, including communication through cursor control and typing \cite{ pandarinath2017high}, %jarosiewicz2015virtual
writing \cite{willett2021high}, and speaking \cite{silva2024speech}. %proix2022imagined, metzger2023high, willett2023high
%These innovations enhance independence and significantly improve quality of life.
%Bidirectional BCIs extend these capabilities by delivering electrical stimulation to the somatosensory cortex to restore sensory feedback, including phantom tactile and proprioceptive sensations \cite{flesher2021brain}. By integrating motor restoration with sensory feedback, these systems create a seamless connection between intent and action, offering a more natural and intuitive user experience.
Recent advancements in BCIs emphasize greater task complexity and dexterity, including high-degree-of-freedom control and  ability to distinguish diverse actions \cite{wodlinger2014ten, silva2024speech}. These systems are paving the way for groundbreaking applications, ranging from cognitive enhancement tools to advanced assistive technologies that adapt to the user’s needs, delivering unprecedented functionality and adaptability \cite{gao2021interface, andersen2022exploring}.

%Beyond control, ML also contributes to improving user adaptability and system reliability in prosthetics. Adaptive algorithms can personalize the decoding models to individual users, accommodating neural signal variability over time. Furthermore, closed-loop feedback systems integrate ML to dynamically adjust device responses based on environmental or user-specific conditions, ensuring seamless interaction.
%These advancements highlight the critical role of ML in empowering neural interfaces to achieve sophisticated, real-time prosthetic functionalities that were previously unattainable.
\vspace{-2mm}
\section{Machine Learning for Neural Interfaces}
%Machine learning (ML) enabling neural interfaces to address a wide range of diagnostic, therapeutic, and assistive challenges. By leveraging large-scale neural recordings, ML algorithms are capable of extracting meaningful patterns and features, which significantly advance the understanding and application of brain signals. This section discusses the critical role of ML in diagnostic/therapeutic and prosthetic applications.

%ML techniques, including both traditional ML models and deep learning (DL) approaches, have revolutionized the analysis of neural data. DL’s ability to process raw neural signals has expanded the scope of neural data analysis, complementing traditional ML methods that rely on pre-defined features. This shift allows ML techniques to uncover hidden neural patterns and brain states more effectively, enhancing the modeling of brain function, interpretation of complex neural dynamics within specific or distributed brain networks, and therefore real-time motor decoding, mental state recognition, and seizure detection. By integrating both feature-based and end-to-end learning approaches, ML-driven neural decoding achieves a balance between accuracy, efficiency, interpretability, and adaptability across diverse neurological applications.

ML techniques, including traditional and deep learning (DL) models, have transformed neural data analysis by uncovering hidden brain patterns while enhancing real-time neural decoding, brain function modeling, and feature-level interpretability.
%ML techniques, including traditional models and deep learning (DL), have transformed neural data analysis by uncovering hidden brain patterns beyond pre-defined features. DL’s ability to process raw neural signals enhances brain function modeling, neural dynamics interpretation, and real-time decoding for motor control, mental state recognition, and seizure detection. By integrating feature-based and end-to-end learning, ML-driven neural decoding balances accuracy, efficiency, interpretability, and adaptability across diverse neurological applications. 
%Traditional ML Intro
%Traditional ML models, such as linear, tree-based, and probabilistic models, have played a crucial role in neural representation learning and decoding [CITE]. Other models, including distance-based techniques and window discrimination provide efficient solutions for decoding neural signals, offering advantages in computational complexity, memory efficiency, real-time processing, and insightful interpretations for neurological applications \cite{Shaeri2022Xform}. These basic methods, such as K-means, classify data based on proximity to class centroids, while window discrimination defines a specific region around each centroid, enabling efficient processing in neural interfaces for tasks like spike detection and sorting \cite{chen2023online, Shaeri2020SFS}.
\vspace{-3mm}
\subsection{Traditioinal Machine Learning}
Traditional ML models, including linear, tree-based, and probabilistic methods, are widely used in neural signal processing, neurological symptom detection and motor decoding. % \cite{guggenmos2013restoration, Shaeri2022Challenges, an2022power, chen2023online, liu2024neural, Shaeri2024MiBMI}. % Shaeri2022Xform %essential for neural representation learning and decoding [CITE].
Distance-based techniques and window discrimination offer computational efficiency, memory savings, and real-time processing, making them well-suited for neural interfaces \cite{shin2022neuraltree, Shaeri2020SFS}. %do2018area
Methods such as K-means and window discrimination classify data by proximity to centroids, shown to be efficient solutions in simple ML tasks, such as spike detection and sorting %for neural interfaces 
\cite{Shaeri2020SFS, chen2023online}. 
% KNN
% K-Nearest Neighbors (KNN) is a non-parametric method highly adaptable to diverse neural signal distributions \cite{peterson2009k}. This flexibility allows KNN to perform well in emotion recognition and early-stage seizure detection where neural activity can be highly individualized \cite{li2018emotion, qing2019interpretable, sharmila2016dwt}. KNN is advantageous for hardware implementation due to its low complexity during training without weight optimization or iterative learning, but its inference can be computationally expensive for large datasets [CITE].
K-Nearest Neighbors (KNN) is a non-parametric, flexible method suited for emotion recognition and early seizure detection, adapting to individualized neural activity \cite{li2018emotion, birjandtalab2017automated}. It offers low-complexity training without weight optimizations, but can be computationally expensive during inference on large datasets \cite{sharmila2016dwt}.

% LDA and SVM
%Linear Discriminant Analysis (LDA) is a linear parametric model widely used for classifying neural signals by maximizing class separability while reducing dimensionality \cite{balakrishnama1998linear}, therefore effectively enhancing motor restoration from neural signals \cite{YS2024JSSC, fu2019improvement}. Since no non-linearity is involved, LDA offers great interpretability with clear decision boundaries while maintaining low computational complexity. Alternatively, Support Vector Machines (SVMs) are effective for decoding neural signals using either a linear setting or non-linear kernels \cite{hearst1998support}. The kernel SVMs are suitable for relatively complex applications such as seizure detection and cognitive state classification \cite{subasi2010eeg, provenza2019decoding}, with efficient inference and low memory usage but high training complexity especially when the number of classes is large [CITE].
Linear Discriminant Analysis (LDA) maximizes class separability while reducing dimensionality, offering an interpretable and low-complexity solution for motor decoding \cite{Shaeri2024MiBMI}. % fu2019improvement, hochberg2012reach
Support Vector Machines (SVMs) effectively decode neural signals, with linear SVMs performing well in simple tasks and kernel SVMs handling complex applications like seizure detection.
%\cite{amin2015feature}. %subasi2010eeg
While SVMs provide efficient inference and low memory usage, the training complexity increases with data size.
% Tree-based: GBDT
% Tree-based models, such as Gradient Boosted Decision Trees (GBDTs) perform well in non-linear and high-dimensional neural data, making them highly effective for classifying cognitive states and detecting neurological disorders \cite{chen2016xgboost}[CITE]. GBDTs rely on hierarchical decision rules, which significantly reduce computational overhead and simplify implementation on hardware. Additionally, their memory usage is relatively low, as only tree structures and split thresholds need to be stored [CITE]. GBDTs offer high interpretability, as feature importance can be extracted for identifying key neural biomarkers linked to neurological disorders [CITE]. GBDT also provides clear decision paths for analyzing decoding outcomes.
Alternatively, Gradient Boosted Decision Trees (GBDTs) excel in non-linear, high-dimensional neural data modeling, making them effective for cognitive state classification and seizure detection\cite{shoaran2018energy, liu2024neural, zhu2020resot}. %zhu2020resot, shoaran2018energy, yao2021predicting
They use hierarchical decision rules, reducing computational overhead and hardware complexity, while requiring minimal memory to store tree structures and split thresholds \cite{taghavi2019hardware}. GBDTs offer high interpretability, enabling neuro-marker identification and providing clear decision paths for neural decoding analysis. 
% Probalisitic models
% Hidden Markov Models (HMMs) and Kalman Filters (KFs) are probabilistic models designed for sequential neural decoding, enhancing performance for motor command prediction and speech decoding \cite{rabiner1986introduction} [CITE]. Both models are hardware-efficient, requiring low computational complexity during inference and minimal memory, making them ideal for on-chip implementations in wearable neuroprosthetics [CITE]. Both HMMs and KFs offer interpretability but are best for discrete and continuous neural/behavioral state tracking respectively [CITE].
Hidden Markov Models (HMMs) and Kalman Filters (KFs) enable sequential neural decoding through probabilistic modeling, iteratively updating hidden states (e.g., motor intentions) by estimating the most probable state based on observed neural data \cite{lorach2023walking, an2022power}. This self-recalibration capability makes them well-suited for real-time neural decoding in adaptive BCIs. %for motor command and speech prediction \cite{rabiner1986introduction} [CITE]. Both are hardware-efficient, with low inference complexity and minimal memory use, making them suitable for on-chip neuroprosthetic applications. HMMs are designed for in discrete state tracking, while KFs handle continuous neural and behavioral state estimation.

\subsection{Deep Learning}
% Summary for Traditional ML and transition to DL
%Collectively, these models provide real-time, energy-efficient, and interpretable solutions for neural interfaces. However, their performance can be limited when dealing with high-dimensional, highly non-linear, and large-scale neural datasets [CITE]. Recently, deep learning (DL) models have emerged as powerful alternatives, capable of automatically learning intricate patterns from raw neural signals [CITE]. DL models overcome these limitations by enabling temporal/spatial, and local/global modeling of neural representations [CITE]. These advances allow DL to surpass traditional methods in scalability and representation learning, making them essential for complex neural decoding and adaptive BCIs [CITE].
Traditional ML models potentially offer real-time and compute-efficient solutions for neural interfaces but typically struggle to capture non-linearities within high-dimensional datasets. % \cite{johnstone2009statistical}.
DL addresses these limitations by automating feature extraction and modeling temporal, spatial, and global neural representations efficiently \cite{willett2021high, silva2024speech, Afzal2024REST}. % song2022eeg, willett2023high, metzger2023high,
DL’s adaptability makes it essential for complex neural decoding and adaptive BCIs, albeit with increased computational complexity.

% Temporal modeling: RNN and CNN
%Recurrent Neural Networks (RNNs) and Convolutional Neural Networks (CNNs) are key for temporal modeling in neural decoding. RNNs capture long-term dependencies, making them effective for speech decoding, seizure detection, and motor command prediction [CITE]. CNNs extract localized temporal patterns through convolutional filters, enabling more efficient processing [CITE]. Unlike RNNs, which require sequential computation and stateful memory, CNNs allow parallelized operations, reducing latency and computational cost [CITE]. Therefore, CNNs are preferable for real-time BCIs and neuroprosthetic control, while RNNs excel in tasks requiring temporal understanding on a longer scale.
Recurrent Neural Networks (RNNs) and Convolutional Neural Networks (CNNs) are essential for temporal modeling in neural decoding \cite{willett2021high, silva2024speech}. RNNs capture long-term dependencies, aiding in motor decoding and seizure detection, %\cite{ahmedt2020neural, tortora2020deep}, %and motor command prediction,
while CNNs extract localized temporal patterns for efficient processing. %\cite{saab2020weak}.
Unlike sequential RNNs, CNNs enable parallelized operations, reducing latency and computational cost for real-time BCIs and neuroprosthetic control, whereas RNNs enhance long-range temporal understanding.
% Spatial modeling: GNN
% Graph Neural Networks (GNNs) model spatial dependencies in neural signals, encoding non-Euclidean geometric adjacency or functional connectivity, making them improve in cognitive state classification and seizure detection [CITE]. However, graph construction and message passing increase the computational load and memory requirements, making GNNs less efficient for low-power, real-time applications [CITE]. However, GNNs can enhance interpretability by mapping brain network topology, providing insights into how different brain regions interact [CITE].
Graph Neural Networks (GNNs) model spatial dependencies in neural signals, advancing mental regulation and seizure classification \cite{zhong2020eeg, Afzal2024REST}. For example, residual state update mechanism (REST) efficiently captures spatiotemporal dependencies in EEG, enabling real-time seizure detection with significantly reduced computation and memory \cite{Afzal2024REST} (Fig. \ref{fig: ML Models}(a)). %GNNs can enhance interpretability by mapping brain network topology, revealing regional interactions. 
% Transformer
% Compared with the other DL models, Transformers overcome long-range dependency challenges in neural decoding by leveraging Multi-Head Self-Attention (MHSA) to model global neural activity patterns [CITE]. This improves seizure detection and emotion recognition, while also enhancing cross-subject generalization in multi-user BCIs [CITE]. Although Transformers demand high computational power and memory due to MHSA, they enable fully parallelized processing, making real-time applications feasible with optimized architectures [CITE]. Additionally, attention maps improve interpretability, highlighting which neural features drive decoding [CITE].
Transformers address long-range dependencies in motor decoding using Multi-Head Self-Attention, enhancing emotion recognition and cross-subject generalization in movement decoding  \cite{jiang2024large, Kalbasi2024DPARS}. %song2022eeg
Though computationally demanding, their parallelized processing enables real-time applications with optimized architectures. Attention maps further improve interpretability by identifying key neural features for decoding.

\subsection{Feature Engineering}
\begin{figure}[!ht]
    \centering    \includegraphics[width=0.89\columnwidth]{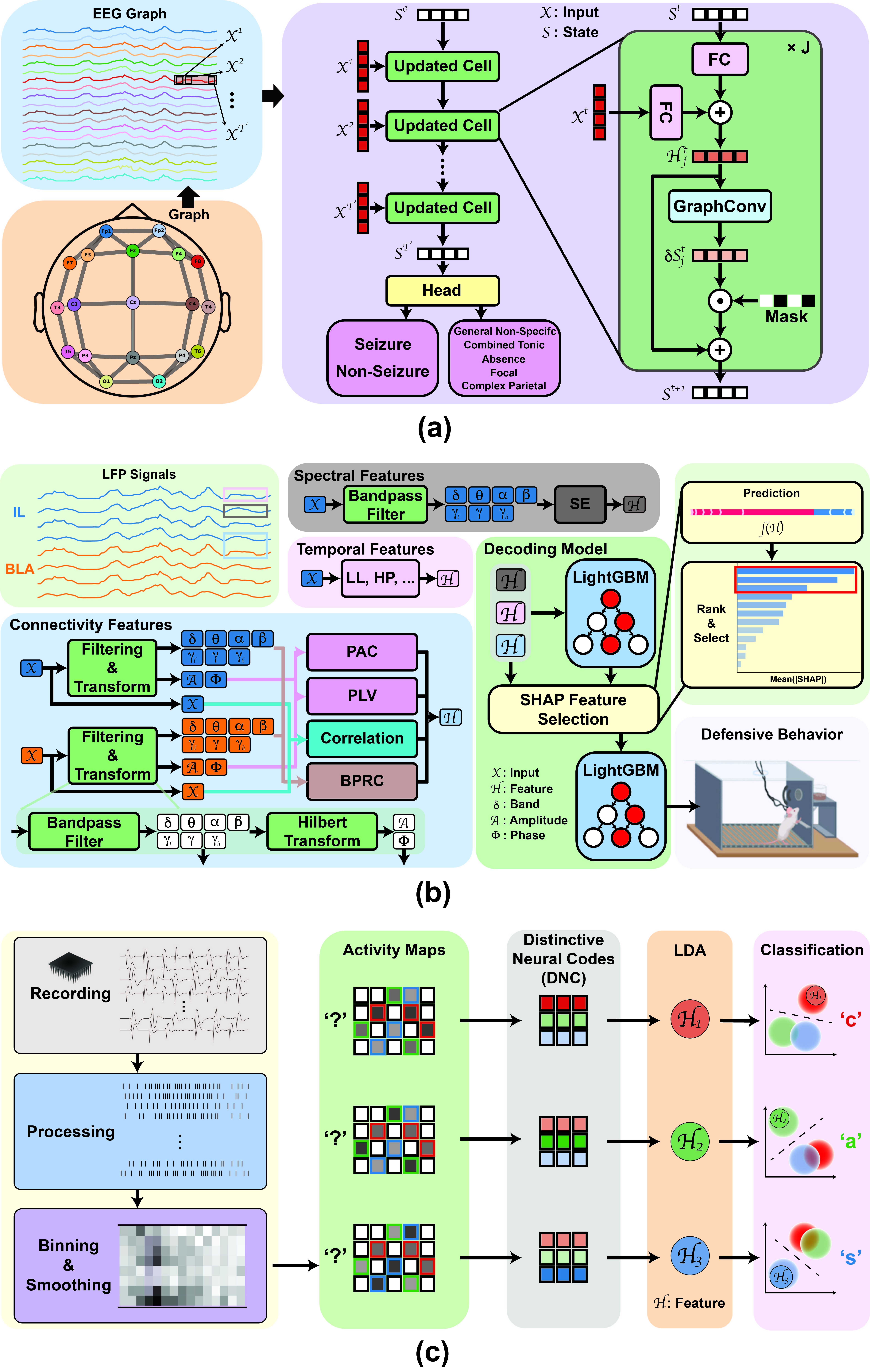} %Fig2.jpg
    \vspace{-11pt}
    \caption{Efficient ML Models for Neural Decoding: (a) Residual State Updates (REST) for seizure detection \cite{Afzal2024REST}. (b) Light Gradient-Boosting Machine (LightGBM) and SHapley Additive exPlanations (SHAP) for decoding anxiety-related behaviors \cite{liu2024neural,zhu2021closed}. (c) Distinctive Neural Code (DNC) and Linear Discriminant Analysis (LDA) for brain-to-text decoding \cite{Shaeri2024MiBMI}.
    }
    \vspace{-13pt}
    \label{fig: ML Models}
\end{figure}

Feature engineering plays a crucial role in neural data analysis (decoding and processing), as it enhances scalability and interpretability. It reduces data dimensionality to enhance computational efficiency while extracting informative features for better interpretation of neural activity.
%On one hand, it reduces data dimensionality, improving computational efficiency and decoding accuracy. On the other hand, extracting informative features facilitates better interpretation of neural activity, enabling more effective and adaptive BCI systems.
%By extracting salient neural features, this process enables more effective classification and real-time processing, making it essential for scalable and adaptive BCI systems.
\begin{comment}
\subsection{Machine Learning Interpretability}
Model interpretability ensures reliable, transparent decision-making, which is crucial for model refinement and justification, especially in high-risk neuroprosthetic and psychiatric applications.
For example, 
\end{comment}
%SHapley Additive exPlanations (SHAP) provide a robust method for quantifying feature importance in neural decoding.
A recent study introduced a lightweight, robust framework that utilizes common neuro-markers—such as spectral energy (SE), line length (LL), phase-amplitude coupling (PAC), phase-locking value (PLV), correlation, and band power ratio between channels (BPRC)—for decoding anxiety-related behaviors, leveraging SHapley Additive exPlanations (SHAP) to quantify feature importance and identify key features \cite{liu2024neural}. SHAP analysis highlighted high-$\gamma$ spectral and connectivity features as key neuro-markers for decoding defensive behaviors in local field potentials (LFPs) recorded from the infralimbic cortex (IL) and basolateral amygdala (BLA) of rats (Fig. \ref{fig: ML Models}(b)). Selecting high-SHAP features preserved decoding accuracy while significantly reducing dimensionality and latency, making the method efficient for real-time, low-power neural decoding in implantable neuropsychiatric systems.
Alternatively, the Distinctive Neural Code (DNC) algorithm employed a class saliency metric to compute feature importance and extract the most distinctive features (Fig. \ref{fig: ML Models}(c)) \cite{YS2024JSSC}. This approach simplified the decoding task, achieving high accuracy with a low-complexity classifier such as LDA.

\section{Neural Systems-on-Chip}
%\st{Implantable next-generation neural devices integrate ML techniques directly on the implant to enable efficient data reduction, minimizing transmission rates and enhancing data privacy by limiting the need to transmit raw brain data. These devices must be highly miniaturized and strictly low-power to ensure safety and suitability for chronic implantation in the brain. These constraints impose significant limitations on available computational resources, requiring aggressive optimization of both hardware and algorithms. Despite these restrictions, such systems must maintain high accuracy, robustness, and low latency to ensure reliability, particularly in life-critical therapeutic and prosthetic applications.}

%\subsection{High-Density Neural Recording SoCs}
Over the past decade, numerous efficient ML and signal processing techniques have been developed, particularly for on-chip spike detection and sorting. For example, adaptive spike detection methods dynamically adjust parameters using adaptive thresholding and nonlinear energy operator (NEO) transformations to enhance sensitivity \cite{razmpour2015signal,guo2022biocas}.
Spike sorting SoCs  incorporate various clustering and classification techniques 
%(e.g., K-means, window discrimination, and decision trees) 
along with feature extraction strategies (e.g., fixed geometric features and salient feature selection) yielding improvements in both accuracy and computational efficiency \cite{Shaeri2020SFS, chen2023online}. %Yang2017AHardware
\begin{comment}
Spike sorting, which classifies detected spikes into distinct neuronal units, employs various clustering and classification techniques, such as K-means, window discrimination, and decision trees. Additionally, feature extraction strategies—including fixed geometric features and salient feature selection are utilized to improve both accuracy and computational efficiency \cite{Yang2017AHardware, Shaeri2020SFS, chen2023online}.
\end{comment}
%$\ell_2$-normalized convolutional autoencoders that optimize feature selection for clustering \cite{Seong2021AMulti}.
%Clustering techniques widely used in spike sorting include K-means and its variants, employing distance metrics such as $\ell_1$-norm\cite{zeinolabedin202216}, $\ell_2$-norm\cite{chen2023online}, cosine similarity\cite{Seong2021AMulti}, or correlation coefficients\cite{kalantari2022hardware} to group similar spike waveforms. 
More advanced neural interface SoCs integrate AI for sophisticated therapeutic and prosthetic applications, enabling real-time adaptation and enhanced functionality. %, particularly in closed-loop neuromodulation and BCI SoCs for clinical and patient-centric applications.
% self-calibration and online adaptability These devices must be highly miniaturized and strictly low-power to ensure safety and long-term stability for chronic implantation in the brain.

\subsection{Neural SoCs for Diagnostic/Therapeutic Applications}
Therapeutic devices must make robust, low-latency decisions while providing real-time feedback (e.g., neurostimulation). Thus, accuracy and latency are particularly critical in applications such as seizure detection and mental regulation \cite{Shoaran2024Intelligent}. 
One such example is a closed-loop device designed to restore motor function following brain injury by re-establishing lost connectivity between cortical areas \cite{guggenmos2013restoration}. It detects premotor cortex action potentials and promptly triggers somatosensory cortex stimulation, facilitating recovery. However, this system lacks scalability and real-time adaptability for dynamically optimizing stimulation parameters.
%\st{While the number of classes is usually low, there are serious challenges in diagnostic and therapeutic applications, particularly with imbalanced datasets [CITE].}
%Local field potential (LFP) data, which captures oscillatory brain activity, provides key neuro-markers such as spectral energy (SE), line length (LL), phase-amplitude coupling (PAC), and phase-locking value (PLV), essential for detecting abnormal neural states. In seizure detection, elevated SE in low-frequency ranges and increased LL characterize preictal activity \cite{esteller2001line}; and for psychiatric disorders, altered PLV in limbic circuits correlate with anxiety-related behaviors and mood dysregulation \cite{likhtik2014prefrontal}. 

\begin{figure}[t]
    \centering    \includegraphics[width=0.95\columnwidth]{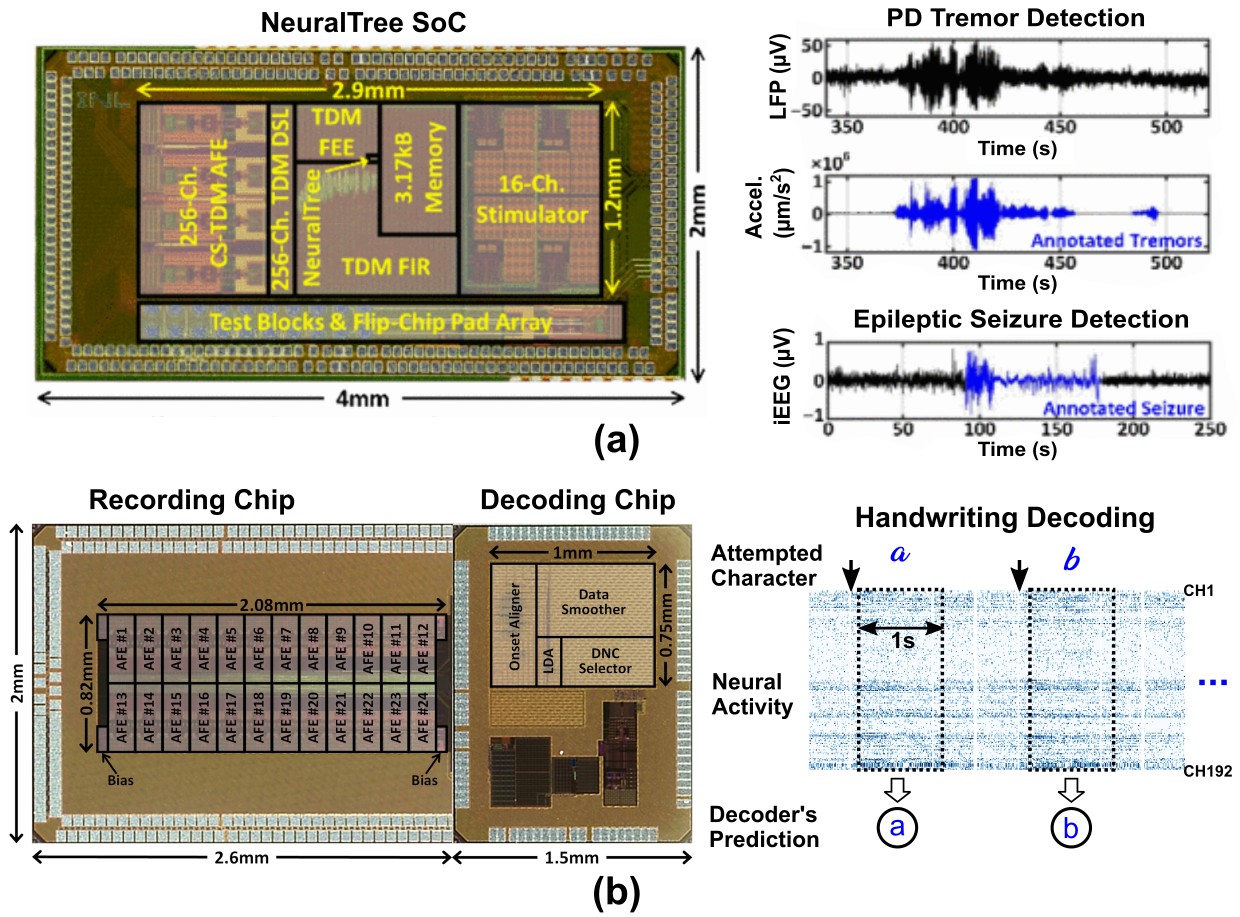} %Fig2.jpg
    \vspace{-10pt}
    \caption{\textbf{Neural SoCs:} (a) Die photo of the closed-loop NeuralTree chip and its experimental results in tremor and seizure detection \cite{shin2022neuraltree}. (b) Die photo of the miniaturized brain-machine interface (MiBMI) chipset and its decoding results in the handwriting task \cite{YS2024JSSC}.}
    \vspace{-10pt}
    \label{fig: Neural SoCs}
\end{figure}

Neural SoCs enhance precision diagnostics and enable targeted interventions for neurological and psychiatric conditions by leveraging key neuro-markers. For example, a neural synchrony processor facilitates precise phase-locked neurostimulation using SE, PAC, and PLV, supporting interventions for anxiety and OCD \cite{shin202216, shin202316}. The NeuralTree SoC integrates versatile neuro-markers with ultra-low-power oblique tree-based classification and multichannel recording/stimulation, enabling real-time adaptive neurotherapies and advancing implantable neural interfaces \cite{shin2022neuraltree} (Fig. \ref{fig: Neural SoCs}(a)). 
%In addition, an EEG-based SoC reinforces SE and LL as seizure neuro-markers, enabling a zero-shot retraining seizure detection processor that adapts in real-time without the need for individualized seizure data \cite{LiuISSCC24AHigh-short}.
%
%Therefore, model interpretability at the neuro-marker level is crucial for life-critical closed-loop neurostimulation systems, ensuring precise and adaptive interventions for clinical applications.% A recent work proposed a lightweight framework for decoding anxiety-related behaviors \cite{liu2024neural}. SHapley Additive exPlanations (SHAP) identified key neuro-markers for decoding defensive behaviors. Selecting high-SHAP features preserved decoding accuracy while significantly reducing memory and latency, making this method efficient for real-time, low-power implantable neuropsychiatric systems.%This recent study emphasizes the role of high-gamma power and inter-regional connectivity in decoding defensive behaviors in psychiatric disorders \cite{liu2024neural}. Using SHapley Additive exPlanations (SHAP), it enhances feature selection transparency, enabling targeted, real-time neuromodulation. This low-latency framework advances precision-driven therapies for anxiety and other neuropsychiatric conditions.
Alternatively, the SciCNN SoC enables patient-independent epilepsy tracking, eliminating the need for pre-deployment retraining and addressing inter-patient seizure variability \cite{tsai2023seizure}. Unlike conventional neuro-marker classifiers, it employs a Seizure-Cluster-Inception CNN on bandpass-filtered EEG/iEEG signals, improving seizure detection in previously unseen patients. However, its computational complexity may pose challenges for low-power implantable applications.

%\cite{tsai2023seizure} % tsai2023scicnn

\subsection{Neural SoCs for Prosthetic Applications}
Recently, BCIs have shown great potential for restoring lost motor capabilities using powerful yet bulky computing systems; however, there is a growing need for implantable or portable neural prostheses that enable seamless use in daily tasks.
%Recently, BCIs have demonstrated high potential for restoring lost motor capabilities using powerful yet bulky benchtop or rackmount computing systems\cite{collinger2013high, willett2021high, willett2023high, metzger2023high}. However, there is an emerging need for implantable/portable neural prostheses that can be easily used by individuals with motor disabilities in daily tasks.
These systems typically utilize high-resolution, high-channel-count neural signals recorded from intracortical Utah or high-density ECoG arrays \cite{patrick2024state}, posing scalability challenges that demand computational and hardware efficiency as channel counts increase while maintaining accuracy \cite{Shaeri2022Challenges}.
Moreover, as neural signals evolve over time due to electrode movement and dynamic brain changes, stable performance requires not only scalable processing but also adaptability via online learning and fast retraining. Consequently, ML SoCs and algorithms must be agile in both training and inference.
%These systems typically utilize high-resolution, high-channel-count neural signals recorded with intracortical Utah arrays or high-density ECoG arrays \cite{patrick2024state}. This raises the challenge of scalability, highlighting the need for computational and hardware efficiency as the number of channels increases tremendously, in addition to maintaining accuracy \cite{Shaeri2022Challenges}. Over time, the data may change due to electrode movement, dynamic brain changes, and other factors. Therefore, such prosthetic devices must be capable of online adaptability and fast retraining. As a result, these ML SoCs and algorithms must be agile in both training and inference.

%However, traditional BCIs rely on large, power-hungry external computers, which limits their portability and practicality for real-world applications. To address these limitations, new approaches in hardware design are focusing on miniaturized, on-chip decoders that can efficiently handle these tasks while reducing size, power consumption, and reliance on external devices.

Following dense neural recording,
these devices extract neural activity features, such as threshold crossings %(multi-unit activity) 
and spiking band power \cite{YS2024JSSC, an2022power} to provide an estimate of neural activity. The neural encoding module then employs ML to capture intricate activity patterns and convert them into actionable commands.
%for movement, writing, and other applications.
So far, only a few studies have explored on-chip decoding for BCI applications due to the challenges of processing high-dimensional data acquired from dense electrodes and the complexity of BCI tasks.
%However, recent advancements have led to the development of neural SoCs capable of decoding complex brain activity.
An early study developed a neuromorphic SoC for decoding four-class neural activity 
%from cortical stimulation 
to control a robotic arm. However, its 16-channel recording capacity limited decoding accuracy and task complexity, while also exhibiting low hardware efficiency \cite{boi2016bidirectional}.
%consumed significant chip area (3.213 mm$^2$/channel) and power (250 $\mu$W/channel), 
Another work used a 128-channel extreme learning machine (ELM) to decode finger movements, implementing the hidden layer on-chip while performing output layer computations on a commercial microcontroller \cite{chen2015128}. However, it still relied on an off-chip processing unit to complete the decoding.
%, consuming 0.0032 $\mu$W/channel and 0.191 mm$^2$/channel \cite{chen2015128}.

A 93-channel intracortical BCI used spiking band power (SBP) features and a steady-state Kalman filter (SSKF) to decode finger movement intentions, but it experienced a latency of up to 2.4 seconds. This system incorporated a commercial Intan analog front-end \cite{an2022power}.
%, with a footprint of 0.097 mm$^2$/channel and power consumption of 6.2 $\mu$W/channel.
Meanwhile, the high-density NeuralTree SoC integrated 256/64-channel ECoG recording and on-chip finger movement decoding using a tree-based neural network. However, it lacked the high-bandwidth spike recording necessary for more complex tasks.
%This system achieved a footprint of 0.02 mm$^2$/channel with a power consumption of 4.24 $\mu$W/channel.
These studies highlight the need for advanced ML-integrated BCI chips capable of handling complex tasks such as handwriting.
%Integrating low-power, custom-designed neural recording units with on-chip decoders can significantly reduce power consumption and miniaturize devices, which is essential for implantable BCIs.

More recently, the miniaturized Brain-to-Text BCI (MiBMI), integrating a neural recording chip with a decoding chip, was introduced (Fig. \ref{fig: Neural SoCs}(b)). It enabled the decoding of intricate motor tasks such as handwriting, using a DNC-based framework combined with LDA.  DNCs reduce the dimensionality of neural data and capture complex  patterns, enabling fast and accurate letter classification with fewer parameters than conventional models. The measured system achieved a software-comparable accuracy in decoding 31 handwritten letters, $\sim 3 \times$ higher in task complexity compared to  state-of-the-art BCI SoCs, with $>10\times$ better area and power efficiency, thanks to the use of DNC algorithm and hardware optimizations (e.g., memory sharing).
%with an ultra-low footprint of 0.0015mm$^2$/channel and a power consumption 0.44$\mu$W/channel.
This chipset presents a promising solution for integrating advanced decoding algorithms into compact, low-power BCIs. %, making it well-suited for long-term implantation in neural prosthetic applications.

\vspace{-1mm}
\section{Conclusion}

Integrating neural interfaces with hardware-optimized ML models on SoC platforms enhances efficiency, speed, and accuracy, aligning with the growing demand for real-time neural decoding and closed-loop neuromodulation in  assistive devices and therapeutic interventions. By leveraging high-density neural recordings and ML-based feature extraction, these systems enable fast, reliable neuro-marker identification, improving applications such as seizure detection, psychiatric treatments, and motor restoration. The synergy between neural interfaces and ML advances miniaturized, energy-efficient SoCs, paving the way for scalable, self-sufficient, and adaptive neural platforms that drive next-generation BCIs, precision medicine, and intelligent neurostimulation systems.

\vspace{-1mm}

% Generated by IEEEtran.bst, version: 1.14 (2015/08/26)

\end{document}